\documentclass[twoside,11pt]{article}

\usepackage{jmlr2e,listings,xcolor,xurl}

\definecolor{codegreen}{rgb}{0,0.6,0}
\definecolor{codegray}{rgb}{0.5,0.5,0.5}
\definecolor{codered}{rgb}{0.58,0,0.0}
\lstdefinestyle{mystyle}{
    commentstyle=\color{codegray},
    keywordstyle=\color{codegreen},
    numberstyle=\tiny\color{codegray},
    stringstyle=\color{codered},
    basicstyle=\ttfamily\footnotesize,
    breakatwhitespace=false,
    breaklines=true,
    captionpos=t,
    keepspaces=true,
    numbers=left,
    numbersep=5pt,
    showspaces=false,
    showstringspaces=false,
    showtabs=false,
    tabsize=2,
    abovecaptionskip=0pt,
    belowcaptionskip=0pt,
}
\jmlrheading{1}{2022}{1-6}{6/3}{10/1}{Cui}{ }

\ShortHeadings{PyTSK: A Python Toolbox for TSK Fuzzy Systems}{PyTSK: A Python Toolbox for TSK Fuzzy Systems}
\firstpageno{1}

\begin{document}

\title{PyTSK: A Python Toolbox for TSK Fuzzy Systems}

\author{\name Yuqi Cui \email yqcui@hust.edu.cn
       \AND
       \name Dongrui Wu\footnote{Corresponding author} \email drwu@hust.edu.cn
       \AND
       \name Xue Jiang \email xuejiang@hust.edu.cn
       \AND
       \name Yifan Xu \email yfxu@hust.edu.cn\\
       \addr  School of Artificial Intelligence and Automation,
       Huazhong University of Science and Technology,       Wuhan, China
       }

\editor{}

\maketitle

\begin{abstract}
This paper presents \texttt{PyTSK}, a Python toolbox for developing Takagi­-Sugeno­-Kang (TSK) fuzzy systems. Based on \texttt{scikit-learn} and \texttt{PyTorch}, \texttt{PyTSK} allows users to optimize TSK fuzzy systems using fuzzy clustering or mini-batch gradient descent (MBGD) based algorithms. Several state-of-the-art MBGD-based optimization algorithms are implemented in the toolbox, which can improve the generalization performance of TSK fuzzy systems, especially for big data applications. \texttt{PyTSK} can also be easily extended and customized for more complicated algorithms, such as modifying the structure of TSK fuzzy systems, developing more sophisticated training algorithms, and combining TSK fuzzy systems with neural networks. The code of \texttt{PyTSK} can be found at \url{https://github.com/YuqiCui/pytsk}.
\end{abstract}


\section{Introduction}

Takagi­-Sugeno­-Kang (TSK) fuzzy systems \cite{Mendel2017} have been widely used in many classification and regression problems \citep{kar2014applications,drwuPL2020}. Traditional TSK fuzzy system optimization algorithms based on fuzzy clustering or evolutionary algorithms are usually time-consuming in big data applications. Mini-batch gradient descent (MBGD) based algorithms \citep{goodfellow2016machine}, which have been widely used in neural network optimization, have also been adapted for TSK fuzzy system optimization in recent years \citep{wu2019optimize,cui2020optimize,du2020tsk,matsumura2017incremental,shi2021fcm}. They have demonstrated better generalization, lower training cost, and better scalability to large datasets. However, few libraries provide convenient and complete Python application programming interfaces (APIs) for developing TSK fuzzy systems, and even fewer based on deep learning frameworks such as \texttt{PyTorch}, \texttt{TensorFlow} and \texttt{Keras}. This may hinder future development and applications of TSK fuzzy systems.

This paper presents \texttt{PyTSK}, a Python toolbox that allows users to develop TSK fuzzy systems using fuzzy clustering or MBGD based optimization algorithms. For fuzzy clustering based optimization, we separate the antecedent and consequent of the TSK fuzzy system so that each part can be easily replaced by a more sophisticated algorithm, and hyper-parameters of both modules can be separately or jointly tuned using APIs in \texttt{scikit-learn}, such as \texttt{GridSearchCV()}. For MBGD based optimization, we provide \texttt{PyTorch}-based APIs for developing TSK fuzzy models. Several state-of-the-art MBGD-based optimization techniques, e.g., uniform regularization \citep{cui2020optimize}, DropRule \citep{wu2019optimize}, and high-dimensional defuzzification \citep{cui2021curse}, are implemented in \texttt{PyTSK} to improve the generalization performance. The API design of \texttt{PyTSK} is similar to that of \texttt{scikit-learn}, i.e., training and prediction of TSK fuzzy models can be performed by calling \texttt{fit()} and \texttt{predict()} functions, respectively.

\section{Design of \texttt{PyTSK}}

This section introduces fuzzy clustering based TSK fuzzy system optimization algorithms, MBGD-based TSK fuzzy system optimization algorithms, and several advanced functionalities in \texttt{PyTSK}.

\subsection{Fuzzy Clustering based TSK Fuzzy System Optimization}

Fuzzy clustering algorithms, e.g., fuzzy c-means \citep{bezdek1984fcm}, can be used to determine the antecedent parameters of a TSK fuzzy system, especially when Gaussian membership functions are used \citep{deng2010scalable}. The input of the consequent part can be transformed into a linear vector, and the consequent parameters can be determined by linear regression algorithms \citep{xu2019concise}.

\texttt{PyTSK} implements the fuzzy c-means algorithm as a subclass of \texttt{TransformerMixin()} in \texttt{scikit-learn}. The \texttt{predict()} method outputs the corresponding membership degree of each cluster, and the \texttt{transform()} method generates the linear consequent input. A complete TSK fuzzy system can be built by using the \texttt{Pipeline()} class of \texttt{scikit-learn} to combine the obtained antecedents with any linear regression algorithm, following \texttt{scikit-learn} API design. A simple example of optimizing a TSK fuzzy classifier using \texttt{PyTSK} is shown in Algorithm~\ref{alg:fcm}.

\lstset{language=Python,frame=lines,caption={Code snippet of optimizing a TSK fuzzy classifier with fuzzy c-means and ridge regression using \texttt{PyTSK} and \texttt{scikit-learn} APIs.},label={alg:fcm}, abovecaptionskip=0pt}
\begin{lstlisting}[float]
from pytsk.cluster import FuzzyCMeans
from sklearn.pipeline import Pipeline
from sklearn.linear_model import RidgeClassifier

model = Pipeline(steps=[
    ('GaussianAntecedent', FuzzyCMeans(n_cluster=n_rule),
    ('Consequent', RidgeClassifier())
])
model.fit(x_train, y_train)  # Model training
y_pred = model.predict(x_test)  # Model prediction
\end{lstlisting}

\subsection{MBGD-based TSK Fuzzy System Optimization}

MBGD-based optimization algorithms are widely used in neural networks, particularly deep learning \citep{goodfellow2016machine}. They have also been adapted for TSK fuzzy system optimization recently \citep{wu2019optimize,cui2020optimize,shi2021fcm}, for better generalization, lower training cost, and better scalability. MBGD-based optimization of fuzzy systems also makes the development of deep fuzzy neural networks easier, as both the fuzzy part and the neural network part can be trained by a single MBGD algorithm.

\begin{figure}[htbp]\centering
  \includegraphics[clip,width=0.95\columnwidth]{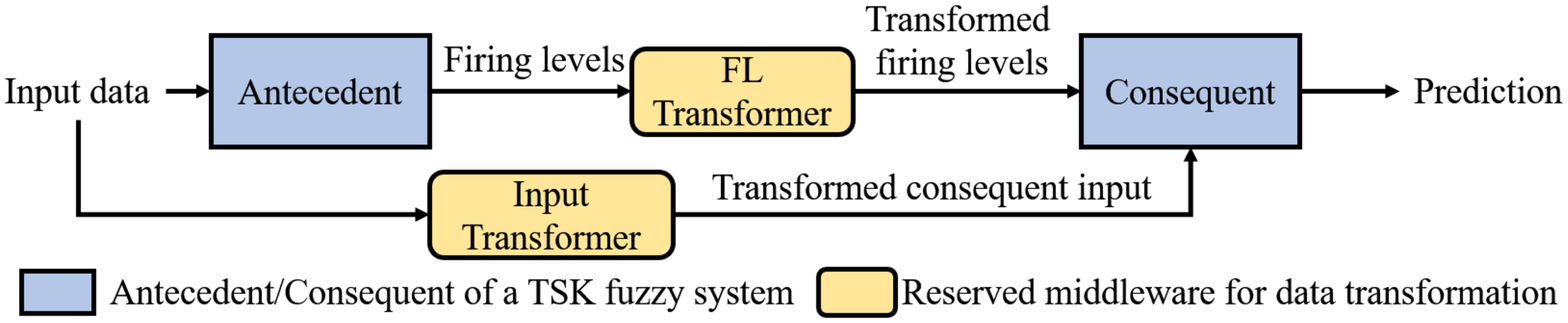}
  \caption{The model structure of a TSK fuzzy system and the reserved middleware in \texttt{PyTSK}.}\label{fig:mbgd}
\end{figure}

In \texttt{PyTSK}, MBGD-based TSK fuzzy system optimization is implemented in \texttt{PyTorch}. Two middleware, Firing Level (FL) Transformer and Input Transformer, are reserved for conveniently modifying the model structure. The model structure of a TSK fuzzy system and the reserved middleware are shown Fig.~\ref{fig:mbgd}.

More specifically, the antecedent module computes the normalized FLs of the rules (\texttt{PyTSK} supports both Gaussian and triangular membership functions), which are then transformed by the FL transformer as needed to impose additional rule weights, normalization, or more transformations. The input transformer transforms the input data as needed and then sends them to the consequent module, which can be used for additional feature selection or normalization.

A simple example of MBGD-based TSK fuzzy classifier optimization using \texttt{PyTSK} is shown in Algorithm~\ref{alg:mbgd}. Note that the defined TSK fuzzy model inherits the \texttt{Module()} class of \texttt{PyTorch}, which means it can be optimized by either our provided API \texttt{Wrapper()} or any other \texttt{PyTorch}-derived API.

\lstset{language=Python,frame=lines,caption={Code snippet of MBGD-based TSK fuzzy classifier optimization using \texttt{PyTSK} and \texttt{PyTorch} APIs.},label={alg:mbgd}}
\begin{lstlisting}[float]
  import torch.nn as nn
  from pytsk.gradient_descent.antecedent import AntecedentGMF
  from pytsk.gradient_descent.training import Wrapper
  from pytsk.gradient_descent.tsk import TSK
  from torch.optim import Adam

  gmf = nn.Sequential(
      AntecedentGMF(in_dim=in_dim, n_rule=n_rule),
      FLTransformer(),
  )  # Defining antecedent and FL transformer
  model = TSK(in_dim=in_dim, out_dim=n_class, n_rule=n_rule, antecedent=gmf, precons=InputTransfomer)  # Define consequent and input transformer
  wrapper = Wrapper(model, optimizer=Adam(model.parameters(), lr=lr), criterion=nn.CrossEntropyLoss())  # Set wrapper
  wrapper.fit(x_train, y_train)  # Model training
  y_pred = wrapper.predict(x_test).argmax(axis=1)  # Model prediction
\end{lstlisting}

\subsection{Advanced Techniques for MBGD-based TSK Fuzzy System Optimization}

Studies have pointed out that MBGD-based TSK fuzzy system optimization may have some limitations. For example, MBGD-based training may generate dead rules and degrade the resulting TSK fuzzy model's fitting ability \citep{cui2020optimize}; the saturation of the softmax function may limit the TSK fuzzy model's performance on high-dimensional datasets \citep{cui2021curse}; the gradient vanishing problem may make TSK fuzzy models sensitive to the choice of the optimizer \citep{cui2022layer}. Therefore, \texttt{PyTSK} supports the following advanced techniques for improving the generalization performance of TSK fuzzy systems\footnote{More examples and details can be found at \url{https://pytskdocs.readthedocs.io/en/latest/models.html}.}:
\begin{enumerate}
	\item \textbf{Uniform regularization} \citep{cui2020optimize}, which forces each rule to have similar contributions to the prediction, so as to avoid dead rules and improve the generalization performance. One can train a TSK fuzzy model with uniform regularization by simply setting the parameter \emph{ur} of the \texttt{Wrapper()} class in \texttt{PyTSK}.

	\item \textbf{High-dimensional TSK (HTSK)} \citep{cui2021curse}, which uses the reciprocal of the input dimensionality as the exponent term of the rule firing levels in defuzzification to improve the generalization performance on high-dimensional datasets. HTSK can be implemented by simply setting the parameter \emph{high\_dim} to True when defining the Gaussian antecedent classes \texttt{AntecedentGMF()} and \texttt{AntecedentShareGM()}.

	\item \textbf{DropRule} \citep{wu2019optimize}, which randomly drops some of the rules during MBGD-based  optimization to improve the generalization performance. DropRule can be implemented by setting the FL Transformer as a \texttt{PyTorch} Dropout \citep{srivastava2014dropout} layer.

	\item \textbf{Deep/stacked fuzzy systems}, which stacks TSK fuzzy systems or deep neural networks to automatically extract more nonlinear features to enhance the model's fitting ability. Since \texttt{PyTSK} implements a TSK fuzzy system as a \texttt{Module()} class of \texttt{PyTorch}, the way to build a deep/stacked TSK fuzzy system is the same as that of building a deep neural network.
\end{enumerate}

\section{Conclusion}

This paper has introduced \texttt{PyTSK}, a Python toolbox for conveniently developing TSK fuzzy systems using fuzzy clustering or MBGD-based optimization algorithms. Models built by \texttt{PyTSK} can be easily customized, and are compatible with other machine learning models derived from \texttt{scikit-learn} and \texttt{PyTorch}. Several state-of-the-art MBGD-based techniques for improving a TSK fuzzy model's generalization performance, e.g., uniform regularization, HTSK and DropRule, have also been implemented in \texttt{PyTSK}.

In the future, we will provide support of more advanced fuzzy clustering algorithms, more types of membership functions, and more regularization approaches for improving the generalization or interpretability of TSK fuzzy systems. \texttt{PyTSK} also welcomes contributions or suggestions from the community.

\acks{This research was supported by the National Natural Science Foundation of China under Grant 61873321, and the Technology Innovation Project of Hubei Province of China under Grant 2019AEA171.}


\begin{thebibliography}{15}
\providecommand{\natexlab}[1]{#1}
\providecommand{\url}[1]{\texttt{#1}}
\expandafter\ifx\csname urlstyle\endcsname\relax
  \providecommand{\doi}[1]{doi: #1}\else
  \providecommand{\doi}{doi: \begingroup \urlstyle{rm}\Url}\fi

\bibitem[Bezdek et~al.(1984)Bezdek, Ehrlich, and Full]{bezdek1984fcm}
James~C. Bezdek, Robert Ehrlich, and William Full.
\newblock {FCM: T}he fuzzy c-means clustering algorithm.
\newblock \emph{Computers \& Geosciences}, 10\penalty0 (2-3):\penalty0
  191--203, 1984.

\bibitem[Cui et~al.(2020)Cui, Wu, and Huang]{cui2020optimize}
Yuqi Cui, Dongrui Wu, and Jian Huang.
\newblock Optimize {TSK} fuzzy systems for classification problems: {M}inibatch
  gradient descent with uniform regularization and batch normalization.
\newblock \emph{IEEE Transactions on Fuzzy Systems}, 28\penalty0 (12):\penalty0
  3065--3075, 2020.

\bibitem[Cui et~al.(2021)Cui, Wu, and Xu]{cui2021curse}
Yuqi Cui, Dongrui Wu, and Yifan Xu.
\newblock Curse of dimensionality for {TSK} fuzzy neural networks:
  {E}xplanation and solutions.
\newblock In \emph{International Joint Conference on Neural Networks}, Virtual
  event, July 2021.

\bibitem[Cui et~al.(2022)Cui, Xu, Peng, and Wu]{cui2022layer}
Yuqi Cui, Yifan Xu, Ruimin Peng, and Dongrui Wu.
\newblock Layer normalization for {TSK} fuzzy system optimization in regression
  problems.
\newblock \emph{IEEE Transcations on Fuzzy Systems}, 2022.
\newblock Under review.

\bibitem[Deng et~al.(2010)Deng, Choi, Chung, and Wang]{deng2010scalable}
Zhaohong Deng, Kup-Sze Choi, Fu-Lai Chung, and Shitong Wang.
\newblock Scalable {TSK} fuzzy modeling for very large datasets using
  minimal-enclosing-ball approximation.
\newblock \emph{IEEE Transactions on Fuzzy Systems}, 19\penalty0 (2):\penalty0
  210--226, 2010.

\bibitem[Du et~al.(2020)Du, Wang, Li, and Liu]{du2020tsk}
Guanglong Du, Zhiyao Wang, Chunquan Li, and Peter~X Liu.
\newblock A {TSK}-type convolutional recurrent fuzzy network for predicting
  driving fatigue.
\newblock \emph{IEEE Transactions on Fuzzy Systems}, 29\penalty0 (8):\penalty0
  2100--2111, 2020.

\bibitem[Goodfellow et~al.(2016)Goodfellow, Bengio, and
  Courville]{goodfellow2016machine}
Ian Goodfellow, Yoshua Bengio, and Aaron Courville.
\newblock \emph{Deep learning}.
\newblock MIT Press, Cambridge, MA, 2016.

\bibitem[Kar et~al.(2014)Kar, Das, and Ghosh]{kar2014applications}
Samarjit Kar, Sujit Das, and Pijush~Kanti Ghosh.
\newblock Applications of neuro fuzzy systems: {A} brief review and future
  outline.
\newblock \emph{Applied Soft Computing}, 15:\penalty0 243--259, 2014.

\bibitem[Matsumura and Nakashima(2017)]{matsumura2017incremental}
Shu Matsumura and Tomoharu Nakashima.
\newblock Incremental learning for {SIRMs} fuzzy systems by {A}dam method.
\newblock In \emph{Joint 17th World Congress of International Fuzzy Systems
  Association and 9th International Conference on Soft Computing and
  Intelligent Systems}, pages 1--4, Otsu, Japan, June 2017.

\bibitem[Mendel(2017)]{Mendel2017}
Jerry~M. Mendel.
\newblock \emph{Uncertain rule-based fuzzy systems: introduction and new
  directions}.
\newblock Springer, 2nd edition, 2017.

\bibitem[Shi et~al.(2021)Shi, Wu, Guo, Zhao, Cui, and Wang]{shi2021fcm}
Zhenhua Shi, Dongrui Wu, Chenfeng Guo, Changming Zhao, Yuqi Cui, and Fei-Yue
  Wang.
\newblock {FCM-RDpA: TSK} fuzzy regression model construction using fuzzy
  {C}-means clustering, regularization, {Droprule, and Powerball Adabelief}.
\newblock \emph{Information Sciences}, 574:\penalty0 490--504, 2021.

\bibitem[Srivastava et~al.(2014)Srivastava, Hinton, Krizhevsky, Sutskever, and
  Salakhutdinov]{srivastava2014dropout}
Nitish Srivastava, Geoffrey Hinton, Alex Krizhevsky, Ilya Sutskever, and Ruslan
  Salakhutdinov.
\newblock Dropout: a simple way to prevent neural networks from overfitting.
\newblock \emph{The Journal of Machine Learning Research}, 15\penalty0
  (1):\penalty0 1929--1958, 2014.

\bibitem[Wu and Mendel(2020)]{drwuPL2020}
Dongrui Wu and Jerry~M. Mendel.
\newblock Patch learning.
\newblock \emph{IEEE Transactions on Fuzzy Systems}, 28\penalty0 (9):\penalty0
  1996--2008, 2020.

\bibitem[Wu et~al.(2019)Wu, Yuan, Huang, and Tan]{wu2019optimize}
Dongrui Wu, Ye~Yuan, Jian Huang, and Yihua Tan.
\newblock Optimize {TSK} fuzzy systems for regression problems: {M}inibatch
  gradient descent with regularization, {DropRule}, and {AdaBound (MBGD-RDA)}.
\newblock \emph{IEEE Transactions on Fuzzy Systems}, 28\penalty0 (5):\penalty0
  1003--1015, 2019.

\bibitem[Xu et~al.(2019)Xu, Deng, Cui, Zhang, Choi, Gu, Wang, and
  Wang]{xu2019concise}
Peng Xu, Zhaohong Deng, Chen Cui, Te~Zhang, Kup-Sze Choi, Suhang Gu, Jun Wang,
  and Shitong Wang.
\newblock Concise fuzzy system modeling integrating soft subspace clustering
  and sparse learning.
\newblock \emph{IEEE Transactions on Fuzzy Systems}, 27\penalty0 (11):\penalty0
  2176--2189, 2019.

\end{thebibliography}

\end{document}